# Direct Processing of Run-Length Compressed Document Image for Segmentation and Characterization of a Specified Block

Mohammed Javed
Dept. of Studies in Computer Science, University of Mysore
Mysore-570006, India
javedsolutions@gmail.com

P. Nagabhushan
Dept. of Studies in Computer Science, University of Mysore
Mysore-570006, India
pnagabhushan@hotmail.com

B.B. Chaudhuri
CVPR Unit
Indian Statistical Institute
Kolkata-700108, India
bbc@isiscal.ac.in

## ABSTRACT

Extracting a block of interest referred to as segmenting a specified block in an image and studying its characteristics is of general research interest, and could be a challenging if such a segmentation task has to be carried out directly in a compressed image. This is the objective of the present research work. The proposal is to evolve a method which would segment and extract a specified block, and carry out its characterization without decompressing a compressed image, for two major reasons that most of the image archives contain images in compressed format and 'decompressing' an image indents additional computing time and space. Specifically in this research work, the proposal is to work on run-length compressed document images.

## 1. INTRODUCTION

Segmentation and characterization of a specified block in a document image finds many applications in the area of Document Image Analysis (DIA) and Pattern Recognition (PR) systems, particularly in applications like signature extraction from official documents [1], logo extraction and detection [17] and document text, photo and line extraction [6]. In addition to this, the concept of block segmentation has been applied for text extraction from layout-aware [18] and 2D plot [12] image documents. Another interesting work uses extracted text blocks from postal images for classification using wavelet coefficients [24]. However, all these methods have major limitation of working only with uncompressed or decompressed documents, although in real life documents are made available in compressed form to provide better transmission and storage efficiency. While dealing with compressed documents, they need to decompress the compressed document and then operate over them. Thus decompression has become an unavoidable prerequisite which indents extra computation time and buffer space. Therefore, it is novel to think of developing intelligent algorithms to extract specified document blocks for document image analysis straight from corresponding compressed formats directly. The specified block to be extracted is conventionally assumed to be rectangular. A sample document showing the real life applications of logo, date and signature extraction by binding them in rectangular blocks is demonstrated in Fig-1.

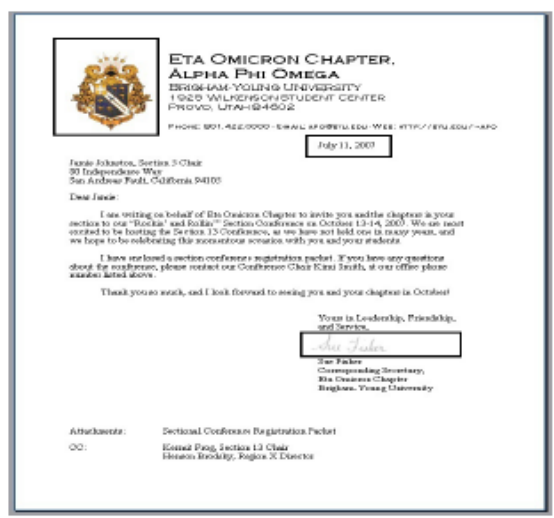

Fig. 1: Block-segments of Logo, Date and Signature used in document image analysis

The second issue is characterization of the segmented/extracted block. Generally absolute characterization of the block may suffice, but there are instances where it is required to express the block characterization contrasted to characterization of the entire document. This is relative characterization of the extracted segment with reference to the entire document. This implies that not only characterization of the extracted block be carried out using the compressed version, also for relative characterization the process has got to be done on the entire document but directly in its compressed version.

Working with compressed version of documents directly for applications in image analysis and pattern recognition, is a challenging goal. The initial idea of working with compressed data was first demonstrated in early 1980's [8, 23]. Run-Length Encoding (RLE), a simple compression method was first used for coding pictures [3] and television signals [14]. There are several efforts in the direction of directly operating on document images in compressed domain. Operations like image rotation [21], connected component extraction [19], skew detection [22], page layout analysis [20] are reported in the literature related to run-length information processing. There are also some initiatives in finding document similarity [13], equivalence [9] and retrieval [15]. One of the recent work using run-length information is to perform morphological related operations [2]. In most of these works, they use either run-length information from the uncompressed image or do some partial decoding to perform the operations. To our best knowledge in the literature, there has been no effort seen in extracting document blocks directly from the TIFF compressed binary text-documents for document characterization. However there is an initiative to show the feasibility of performing segmentation without decompression using JPEG documents by [4, 5], which may not be suitable for direct processing of TIFF compressed documents.

Therefore in this backdrop, a novel idea of segmenting a document block straight from the run-length data of TIFF compressed documents, without going through the stage of decompression is proposed. Some of the recent works related to feature extraction and text-document segmentation directly from run-length compressed data can be found in [11, 10, 16]. Rest of the paper is organized as follows: section-2 describes the problem, related issues and terminologies, section-3 discusses the proposed model for segmenting a block, section-4 shows the experimental analysis with the proposed methods, section-5 brings out absolute and relative characterization of the segmented block using density and entropy values and finally section-6 concludes the paper with a brief summary.

## 2. UNDERSTANDING THE PROBLEM AND TERMINOLOGIES

Segmenting and analyzing or characterizing a specified block from documents is more frequently used in document image analysis, which usually means extracting the contents of a document bound in some rectangular segment. Such a segmentation involves extraction of horizontal and vertical boundaries which can be performed easily in an decompressed image without much effort using the specified rows and columns of the block as shown in Fig-2a. However, when this image is subjected to run-length compression, the specified block does not visibly appear rectangular shaped. Therefore automatically tracing the block becomes challenging. While tracing the contents of the specified block of Fig-2a within the compressed data, it is observed to be located within the enclosed boundary of runs shown in Fig-2b. This boundary can be located inside the compressed data with the help of start (y1) and end (y2) columns of the specified block. For example, at line number 5 of the Fig-2b, the start-column (y1 = 3) and end-column (y2 = 6) of the specified block of Fig-2a is located within the run columns of 1 (start-run) and 2 (end-run) respectively and hence, the position of the specified block inside the compressed data is tabulated respectively as P1 and P2 in Fig-2b. Once the boundary of the specified block is approximately located inside the compressed data as in Fig-2b, it is necessary to do further refinement of the runs specifically at the runs of start and end locations of the block to get exact boundary of the specified block as in Fig-2c. The schematic view of these exact locations of the specified block in the compressed data is given in Fig-2d.

Now let us try to understand the extraction process of exact boundaries of the specified block using the position table shown in Fig-2b. For example, consider the block boundary runs pointed by P1 and P2 at line number 5 of Fig-2b which is schematically shown in Fig-3. It is observed that the start and end boundaries of the specified block is located in between the runs pointed by P1(4) and P2(4). Therefore, the extraction of exact block warrants the removal of excess runs from the both ends. These extra runs at both ends are coined as residue runs which are correspondingly tabulated as R1(2) and R2(2) in Fig-2b. These residue runs have to be eliminated from the enclosed boundary of specified block in compressed data of Fig-2b and the updated compressed data is shown in Fig-4a. The resulting compressed data when decompressed gives the exact specified block as shown in Fig-4b.

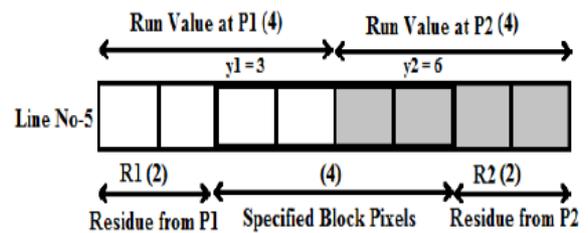

**Fig. 3: Specified block runs and their corresponding residue runs**

Another important aspect in this research work is the absolute and relative characterization of the specified block from the compressed data. The method of analyzing the contents of the segmented block independently is absolute characterization, whereas the analysis with respect to source document is relative characterization. In this research study, the entropy [16, 7] and density features are used for characterizing the specified block which will be discussed in section-5.

## 3. PROPOSED MODEL

In this section, a novel method for extracting specified document blocks straight from run-length compressed document data is proposed. The different stages involved in segmenting a document block are shown in Fig-5.

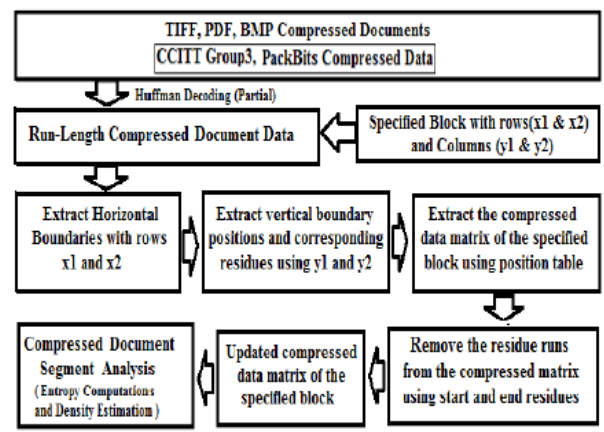

**Fig. 5: Proposed model for block segmentation from compressed data**

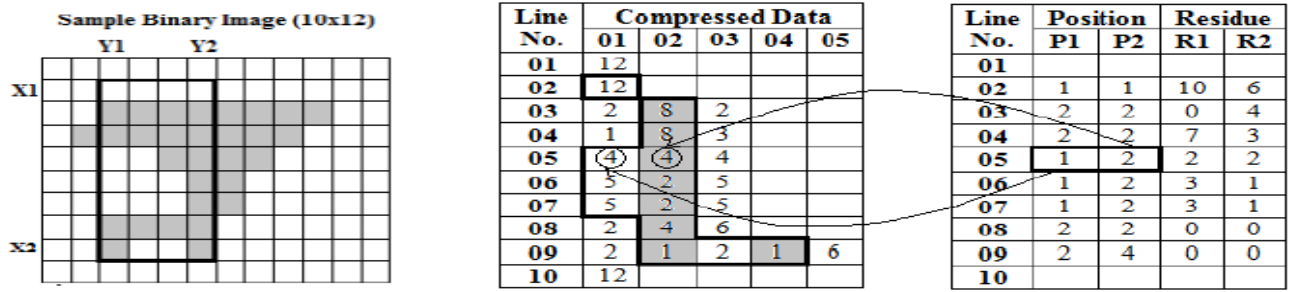

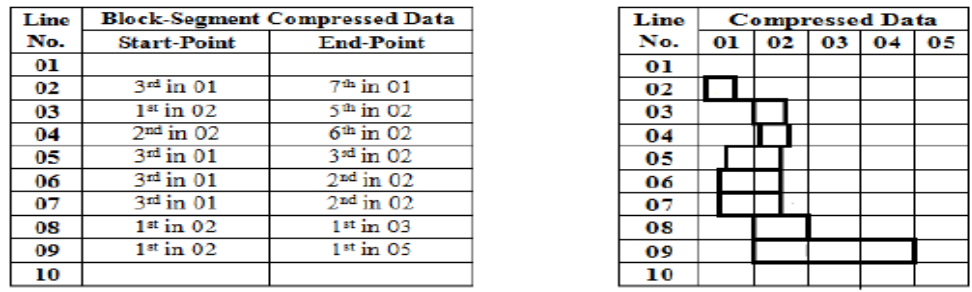

Fig. 2: Schematic view of block-segment in compressed and decompressed binary image

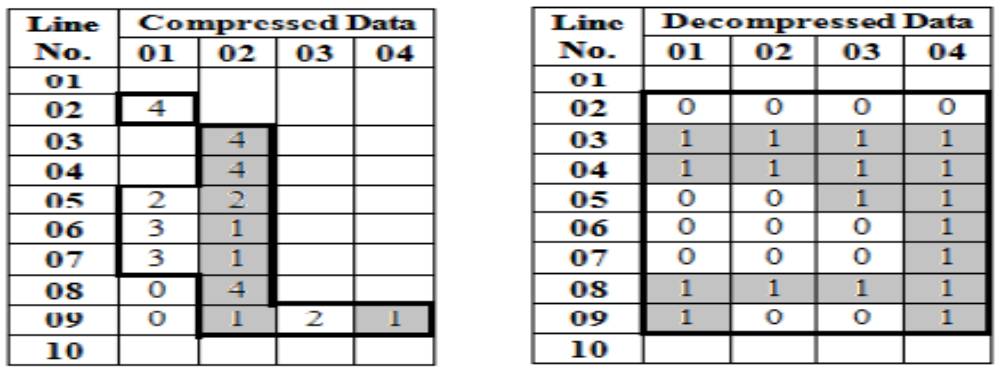

Fig. 4: Compressed and decompressed version of the specified block segment

The proposed model works on the run-length compressed data extracted using Huffman decoding from the TIFF compressed binary documents. Using the specified rows (x1 and x2) of the document block, the segmentation of horizontal boundaries is done in the compressed data. The next stage is to trace the vertical boundary positions and corresponding residue runs from each row in the compressed data using columns (y1 and y2) of the specified block. The strategy to be used in locating the boundary position and the corresponding residues is summarized in Table-1 and Table-2, where 'j' indicates the column position of runs in the compressed data. For every row of runs in compressed data, the cumulative sum of runs (runSum) is calculated until one of the following cases described in Table-1 is encountered, which gives the position of y1 in the compressed data. Similarly continue the row scan to find the position of y2 based on the condition specified in Table-2. Tabulate these positions of y1 and y2 traced from the compressed data as P1 and P2 and their corresponding residues R1 and R2 in position table shown in Fig-2b. Now using the position table extract the compressed data matrix of the specified block. As it is known that, this compressed data approximately represents the specified block and hence requires removal of residue runs using the residue values tabulated previously. Once the residue runs are removed from the compressed data of the specified block using the strategy specified in Table-3, the updated compressed data is obtained which is shown with an example in Fig-4a. The resulting compressed data of the specified block is utilized for further document analysis such as for absolute and relative characterization using the density and entropy features which will be discussed in the next section.

**Table 3. : Strategy for residue run modifications from the compressed data**

| For all the rows 'i' | Updated Start Run | Updated End Run |
|---|---|---|
| Case 1: $if P1 = P2 \ \& \ R1 \neq 0 \ \& \ R2 \neq 0$ | $M(i, P1) = R1 - R2$ | |
| Case 2: $if P1 = P2 \ \& \ R1 \neq 0 \ \& \ R2 = 0$ | $M(i, P1) = M(i, P1) - R1$ | |
| Case 3: $if P1 = P2 \ \& \ R1 = 0 \ \& \ R2 \neq 0$ | $M(i, P1) = M(i, P2) - R2$ | |
| Case 4: $if P1 = P2 \ \& \ R1 = 0 \ \& \ R2 = 0$ | $M(i, P1) = M(i, P1)$ | |
| Case 5: $if P1 \neq P2 \ \& \ R1 = 0 \ \& \ R2 = 0$ | $M(i, P1) = M(i, P1)$ | $M(i, P2) = M(i, P2)$ |
| Case 6: $if P1 \neq P2 \ \& \ R1 = 0 \ \& \ R2 \neq 0$ | $M(i, P1) = M(i, P1)$ | $M(i, P2) = M(i, P2) - R2$ |
| Case 7: $if P1 \neq P2 \ \& \ R1 \neq 0 \ \& \ R2 = 0$ | $M(i, P1) = R1$ | $M(i, P2) = M(i, P2)$ |
| Case 8: $if P1 \neq P2 \ \& \ R1 \neq 0 \ \& \ R2 \neq 0$ | $M(i, P1) = R1$ | $M(i, P2) = M(i, P2) - R2$ |

**Table 1. : Locating the position and corresponding residue of vertical start boundary (y1) of the specified block in the compressed data**

| | Start Position (P1) | Residue Run (R1) |
|---|---|---|
| Case 1: $runSum > y1$ | $P1 = j$ | $R1 = runSum - y1 + 1$ |
| Case 2: $runSum = y1 - 1$ | $P1 = j + 1$ | $R1 = 0$ |

**Table 2. : Locating the position and corresponding residue of vertical end boundary (y2) of the specified block in the compressed data**

| | End Position (P2) | Residue Run (R2) |
|---|---|---|
| Case 1: $runSum > y2$ | $P2 = j$ | $R2 = runSum - y2$ |
| Case 2: $runSum = y2$ | $P2 = j$ | $R2 = 0$ |

## 4. EXPERIMENTAL ANALYSIS

In this section, the extraction of document blocks directly from the run-length compressed data is experimentally demonstrated. The experimental results of working on compressed data are difficult to present and visualize. However in this research study, for the better visualization and understanding of the proposed methods and results, the resulting compressed data of extracted segments is decompressed and shown in this section. Consider a decompressed version of a sample document of size 1009 X 1542 shown in Fig-6a. The decompressed version of four segmented document blocks from the sample document with respective rows and columns are shown in Fig-6b (300 _ 300, where x1 = 100; x2 = 400; y1 = 200; y2 = 500), Fig-6c (300 _ 400, where x1 = 500; x2 = 800; y1 = 700; y2 = 1100), Fig-6d (300 _ 300, where x1 = 700; x2 = 1000; y1 = 1200; y2 = 1500) and Fig-6e (400 _ 300, where x1 = 100; x2 = 500; y1 = 1200; y2 = 1500). Similarly the proposed method has been tested randomly with 35 compressed documents from Bengali, Kannada and English scripts.

The performance of the results obtained from the proposed method can be evaluated in two ways. One method of finding accuracy of the segmented block is to decompress the compressed data of the block $A_{i,j}$ and match it with that of uncompressed version of ground truth block $B_{i,j}$ using the formula,

$$Accuracy(\%) = \left[1 - \frac{\sum_{i=1}^{m}\sum_{j=1}^{n}|A_{i,j}-B_{i,j}|}{m \times n}\right] \times 100$$

where m and n are number of rows and columns in the given ground truth block segment and decompressed version of extracted block. The second method of calculating accuracy is to have a ground truth of compressed data of block $B_{i,j}$ and compare it with the compressed data of extracted block $A_{i,j}$ with the following formula,

$$Accuracy(\%) = \left[1 - \frac{\sum_{i=1}^{m}\sum_{j=1}^{n'}\{|A_{i,j}-B_{i,j}|\}}{m \times \sum_{j=1}^{n'} A_{1,j}}\right] \times 100$$

Where m and n' are the rows and columns of the specified block in the compressed data, where n'<n. The accuracy of 4 segmented blocks in Fig-6 using both the methods described is tabulated in Table-4.

**Table 4. : Accuracy of the extracted blocks**

| Sample | Accuracy-1 (%) | Accuracy-2 (%) |
|---|---|---|
| Block-Segment-1 | 100 | 100 |
| Block-Segment-2 | 100 | 100 |
| Block-Segment-3 | 100 | 100 |
| Block-Segment-4 | 100 | 100 |

## 5. BLOCK CHARACTERIZATION

In this section, the absolute and relative characterization of the specified block extracted directly from the compressed data is demonstrated. The density and entropy features [11, 16, 7] computed from Conventional Entropy Quantifiers (CEQ) and Sequential Entropy Quantifiers (SEQ) are used for characterizing the extracted document block. CEQ measures the energy contribution of each row by considering the probable occurrence of +ve and –ve transitions among the total number of pixels in that row and SEQ analyzes the component by measuring the entropy at the position of occurrence of the transitions. The detailed presentation is available in [7].

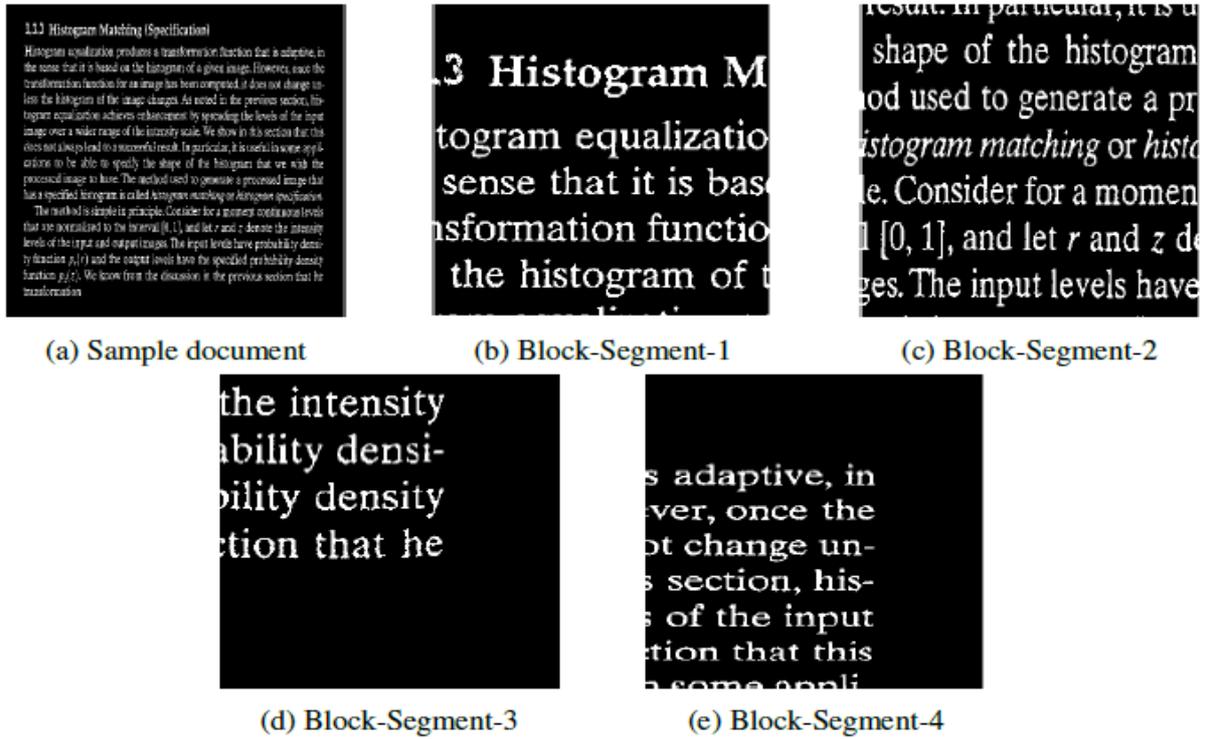

**Fig. 6: Viewing the sample document and different block-segments extracted from compressed data after decompression**

The mathematical formula for CEQ is as follows,

$$E(t) = p * \log(\tfrac{1}{p}) + (1 - p) * \log(\tfrac{1}{(1-p)})$$

where t is the transition from 0 - 1 and 1 - 0, E(t) is the entropy, p is the probable occurrence of transition in each row, then 1 - p is the probable non-occurrence of transition. For SEQ, if the transition occurs between two columns $C_\beta 1$ and $C_\beta 2$ in row $r_\alpha$ then corresponding row entropy is formulated as:

$$E(\beta) = \tfrac{r_\alpha}{m}\left(\tfrac{pos}{n} * \log \tfrac{n}{pos} + (m - \tfrac{pos}{n}) * \log \tfrac{m}{m*n-pos}\right)$$

where β = 1,…,m and the position parameter pos, which indicates the position of transition point in horizontal direction.

The absolute characterization of the extracted block is done considering only the features of the block. The parameters 'p' for CEQ and SEQ is computed as the ratio of total number of (0-1) Or (1- 0) transitions in each row of the block and total number of probable transitions possible in each row of the block, and parameter 'pos' indicates the column position of transition in each row of the block. On the other hand, the relative characterization of the specified block is done taking into account the features of source document. The parameters 'p' for CEQ and SEQ is computed as the ratio of total number of (0-1) Or (1-0) transitions in each row of the block and total number of probable transitions possible in a row of block with respect to source document, and the parameter 'pos' indicates the column position of transition in a row of block with respect to source document.

On the similar line, the absolute and relative characterization using density feature is given as follows,

$$\text{Absolute Density} = \frac{\text{Total sum of foreground pixel runs}}{\text{Size of the block}}$$

$$\text{Relative Density} = \frac{\text{Total sum of foreground pixel runs}}{\text{Size of the Document}}$$

The absolute and relative characterization using density and entropy features for different blocks extracted in Fig-6 is tabulated in Tables-5 and Table-6 respectively. From these tables it is observed that the extracted block in Fig-6c is a case for high density and high entropy, Fig-6d is an example for low density and low entropy. On the other hand Fig-6e shows low density and high entropy and Fig-6b shows high density and low entropy.

**Table 5. : Absolute Density and Entropy computations of sample document and extracted blocks**

| Sample | Density | CEQ | SEQ |
|---|---|---|---|
| Sample Document | $D = 0.0945$ | $C = 75.321$ | $S = -1.096 \times 10^8$ |
| Block-Segment-1 | $D1 = 0.1326$ | $C1 = 27.873$ | $S1 = -2.217 \times 10^6$ |
| Block-Segment-2 | $D2 = 0.1556$ | $C2 = 35.070$ | $S2 = -3.117 \times 10^6$ |
| Block-Segment-3 | $D3 = 0.0677$ | $C3 = 17.183$ | $S3 = -6.886 \times 10^5$ |
| Block-Segment-4 | $D4 = 0.0797$ | $C4 = 25.514$ | $S4 = -2.860 \times 10^6$ |

**Table 6. : Relative Density and Entropy computations of sample document and extracted blocks**

| Sample | Density | CEQ | SEQ |
|---|---|---|---|
| Sample Document | $D = 0.0945$ | $C = 75.321$ | $S = -1.096 \times 10^8$ |
| Block-Segment-1 | $D1 = 0.0077$ | $C1 = 7.899$ | $S1 = -4.260 \times 10^6$ |
| Block-Segment-2 | $D2 = 0.0121$ | $C2 = 12.560$ | $S2 = -1.690 \times 10^7$ |
| Block-Segment-3 | $D3 = 0.0039$ | $C3 = 4.721$ | $S3 = -6.953 \times 10^6$ |
| Block-Segment-4 | $D4 = 0.0062$ | $C4 = 7.045$ | $S4 = -4.635 \times 10^6$ |

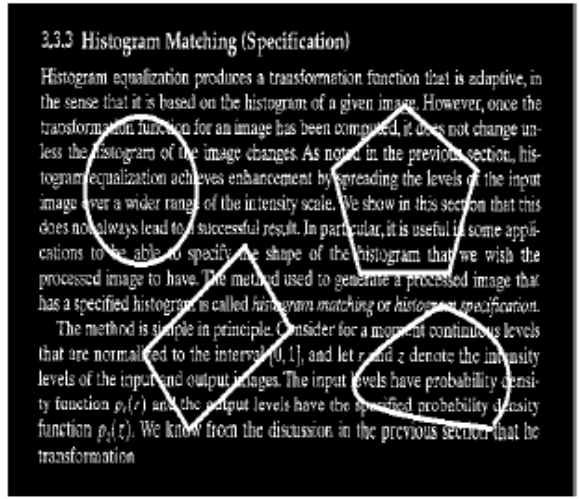

**Fig. 7: Extracting document segments of various shapes**

In this research work, the study is limited to extracting and characterizing the document blocks in rectangular segments from the compressed data. However, there is a scope for research in extracting document blocks of different shapes and also considering skew as shown in Fig-7, which could be taken as an extension to this research work.

## 6. CONCLUSION

In this research work, a novel method for extracting the specified document block in rectangular segments directly from the run length compressed data without going through the stage of decompression is proposed. Further, the absolute and relative characterization of the extracted blocks using density and entropy features are demonstrated. This research study also demonstrates that document analysis is possible using compressed data without decompression and also opens up the gateway for plenty of research issues using compressed documents in compressed domain.

## 2. REFERENCES

[1] Sheraz Ahmed, Muhammad Imran Malik, Marcus Liwicki, and Andreas Dengel. Signature segmentation from document images. International Conference on Frontiers in Handwriting Recognition (ICFHR), pages 423–427, 2012.

[2] Thomas M. Breuel. Binary morphology and related operationson run-length representations. International Conferenceon Computer Vision Theory and Applications - VISAPP, pages 159–166, 2008.

[3] J. Capon. A probabilistic model for run-length coding of pictures.IRE Transactions on Information Theory, 5:157–163,1959.

[4] Ricardo L. de Queiroz and Reiner Eschbach. Segmentation of compressed documents. Proceedings of International Conference on Image Processing, 3:70–73, 1997.

[5] Ricardo L. de Queiroz and Reiner Eschbach. Fast segmentation of the jpeg compressed documents. Journal of Electronic Imaging, 7(2):367–377, 1998.

[6] M. Sezer Erkilinc, Mustafa Jaber, Eli Saber, Peter Bauer, and Dejan Depalov. Text, photo, and line extraction in scanneddocuments. Journal of Electronic Imaging, 21(3):033006–1–033006–18, 2012.

[7] Sahana D. Gowda and P Nagabhushan. Entropy quantifiers useful for establishing equivalence between text document images. International Conference on Computational Intelligence and Multimedia Applications, pages 420 – 425, 2007.

[8] G. Grant and A.F. Reid. An efficient algorithm for boundarytracing and feature extraction. Computer Graphics and Image Processing, 17:225–237, November 1981.

[9] Jonathan J. Hull. Document image similarity and equivalence det. International Journal on Document Analysis and Recognition (IJDAR'98), 1:37–42, 1998.

[10] Mohammed Javed, P Nagabhushan, and B B Chaudhuri. Extraction of line-word-character segments directly from runlength compressed printed text-documents. National Conference on Computer Vision, Pattern Recognition, Image Processing and Graphics (NCVPRIPG'13), Jodhpur, India, December19-21, 2013 in Press.

[11] Mohammed Javed, P Nagabhushan, and B B Chaudhuri. Extraction of projection profile, run-histogram and entropy features straight from run-length compressed documents. Proceedings of Second IAPR Asian Conference on Pattern Recognition (ACPR'13), Okinawa, Japan, November 2013.

[12] Saurabh Kataria, William Browuer, Prasenjit Mitra, and C. Lee Giles. Automatic extraction of data points and text blocks from 2-dimensional plots in digital documents. Association for the Advancement of Artificial Intelligence, 2008.

[13] Dar Shyang Lee and Jonathan J. Hull. Detecting duplicates among symbolically compressed images in a large document database. Pattern Recognition Letters, 22:545–550, 2001.

[14] J.O. Limb and I.G. Sutherland. Run-length coding of television signals. Proceedings of IEEE, 53:169–170, 1965.

[15] Yue Lu and Chew Lim Tan. Document retrieval from compressed images. Pattern Recognition, 36:987–996, 2003.

[16] P.Nagabhushan, Mohammed Javed, and B.B.Chaudhuri. Entropy computation of document images in run-length compressed domain. International Conference on Signal and Image Processing (ICSIP14), Bangalore, India, January 8-11, 2014 in Press.

[17] Arash Asef Nejad and Karim Faez. A novel method for extracting and recognizing logos. International Journal of Electrical and Computer Engineering (IJECE), 2(5):577–588, October 2012.

[18] Cartic Ramakrishnan, Abhishek Patnia, Eduard Hovy, and Gully APC Burns. Layout-aware text extraction from fulltext pdf of scientific articles. Source Code for Biology and Medicine, 7:7, 2012.

[19] E. Regentova, S. Latifi, S. Deng, and D. Yao. An algorithm with reduced operations for connected components detection in itu-t group 3/4 coded images. IEEE Transactions on Pattern Analysis and Machine Intelligence, 24(8):1039 – 1047, August 2002.

[20] E.E. Regentova, S. Latifi, D. Chen, K. Taghva, and D. Yao. Document analysis by processing jbig-encoded images. International Journal on Document Analysis and Recognition (IJDAR), 7:260–272, 2005.

[21] Yoshihiro Shima, Seiji Kashioka, and Jun'Ichi Higashino. A high-speed algorithm for propagation-type labeling based on block sorting of runs in binary images. Proceedings of $10^{th}$ International Conference on Pattern Recognition (ICPR), 1:655–658, 1990.

[22] A. Lawrence Spitz. Analysis of compressed document images for dominant skew, multiple skew, and logotype detection. Computer vision and Image Understanding, 70(3):321–334, June 1998.

[23] T. Tsuiki, T. Aoki, and S. Kino. Image processing based on a runlength coding and its application to an intelligent facsimile. Proc. Conf. Record, GLOBECOM '82, pages B6.5.1–B6.5.7, November 1982.

[24] K R Varshney. Block-segmentation and classification of grayscale postal images. Report in School of Electrical and Computer Engineering, Cornell University, 2004.